\documentclass{article}

\usepackage{times}
\usepackage{graphicx} 
\usepackage{subfigure} 
\usepackage{todonotes}

\usepackage{algorithm}
\usepackage{algorithmic}

\usepackage[accepted,nohyperref]{icml2015} 

\usepackage{tikz}
\usepackage{pgfplots}
\usetikzlibrary{arrows,backgrounds,fit,positioning,pgfplots.groupplots}
\tikzstyle{int}=[draw, minimum size=1em]
\tikzstyle{init} = [pin edge={to-,thin,black}]
\tikzstyle{surround} = [draw, minimum size=2em]
\usepackage[framemethod=TikZ]{mdframed}
\usepackage{comment}
\usepackage{amsmath}
\usepackage{amssymb}
\usepackage{mynotation}
\usepackage{siunitx}

\newcommand{\figspace}{}

\newcommand{\myargmin}[2]{\underset{#1}{\mathrm{arg}\,\mathrm{min}}\,\,#2}
\newcommand{\feature}{z}

\newcommand{\meas}{y}

\newcommand{\tind}{t}
\newcommand{\tend}{N}

\newcommand{\ndim}{n}

\newcommand{\hn}{h_n}

\newcommand{\measdim}{M}
\newcommand{\featuredim}{m}

\newcommand{\unaryminus}{\scalebox{0.75}[1.0]{\( - \)}}

\newcommand{\pilcofile}{data/pilcoAE_wrap.dat}

\icmltitlerunning{From Pixels to Torques: Policy Learning with Deep Dynamical Models}

\begin{document} 

\twocolumn[
\icmltitle{From Pixels to Torques: Policy Learning with Deep Dynamical Models}



\icmlauthor{Niklas Wahlstr\"om}{nikwa@isy.liu.se}
\icmladdress{Division of Automatic Control, Link\"oping University, Link\"oping, Sweden}
\icmlauthor{Thomas B. Sch\"on}{thomas.schon@it.uu.se}
\icmladdress{Department of Information Technology, Uppsala University, Sweden}
\icmlauthor{Marc Peter Deisenroth}{m.deisenroth@imperial.ac.uk}
\icmladdress{Department of Computing, Imperial College London, United Kingdom}

\icmlkeywords{Deep neural networks, nonlinear systems, low-dimensional embedding, auto-encoder, model predictive control}

\vskip 0.3in
]

\begin{abstract} 
Data-efficient learning in continuous state-action spaces using very high-dimensional observations remains a key challenge in developing fully autonomous systems. In this paper, we consider one instance of this challenge, the pixels to torques problem, where an agent must learn a closed-loop control policy from pixel information only. We introduce a data-efficient, model-based reinforcement learning algorithm that learns such a closed-loop policy directly from pixel information. The key ingredient is a deep dynamical model that uses deep auto-encoders to learn a low-dimensional embedding of images jointly with a predictive model in this low-dimensional feature space. Joint learning ensures that not only static but also dynamic properties of the data are accounted for. This is crucial for long-term predictions, which lie at the core of the adaptive model predictive control strategy that we use for closed-loop control. Compared to state-of-the-art reinforcement learning methods for continuous states and actions, our approach learns quickly, scales to high-dimensional state spaces and is an important step toward fully autonomous learning from pixels to torques.
\end{abstract} 

\section{Introduction}
\label{sec:introduction}

The vision of fully autonomous and intelligent systems that learn by themselves has influenced AI and robotics research for many decades. To devise fully autonomous systems, it is necessary to (1) process perceptual data (e.g., images) to summarize knowledge about the surrounding environment and the system's behavior in this environment, (2) make decisions based on uncertain and incomplete information, (3) take new information into account for learning and adaptation. Effectively, any fully autonomous system has to close this perception-action-learning loop without relying on specific human expert knowledge.
The \emph{pixels to torques problem}~\cite{Brock2011} identifies key aspects of an autonomous system: autonomous thinking and decision making using sensor measurements only, intelligent exploration and learning from mistakes. 

We consider the problem of learning closed-loop policies (``torques'') from pixel information end-to-end. A possible scenario is a scene in which a robot is moving about. The only available sensor information is provided by a camera, i.e., no direct information of the robot's joint configuration is available. The objective is to learn a continuous-valued policy that allows the robotic agent to solve a task in this continuous environment in a data-efficient way, i.e., we want to keep the number of trials small.
To date, there is no fully autonomous system that convincingly closes the perception-action-learning loop and solves the pixels to torques problem in continuous state-action spaces, the natural domains in robotics. 

\begin{figure*}[t]
\centering
\includegraphics[width = \hsize]{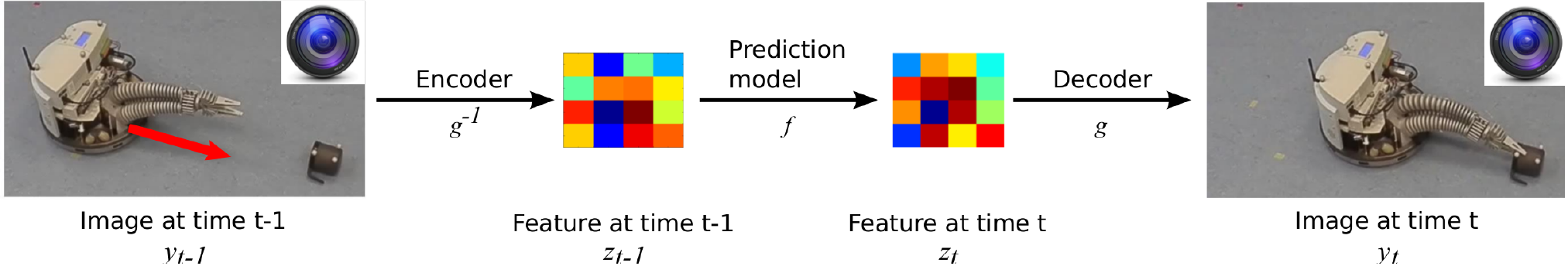}
\caption{Illustration of our idea of combining deep learning architectures for feature learning and prediction models in feature space. A camera observes a robot approaching an object. A good low-dimensional feature representation of an image is important for learning a predictive model if the camera is the only sensor available.}
\label{fig:model_illustration}
\end{figure*}

A promising approach toward solving the pixels to torques problem is Reinforcement Learning (RL) \cite{Sutton1998}, a principled mathematical framework that deals with fully autonomous learning from trial and error. However, one practical shortcoming of many existing RL algorithms is that they require many trials to learn good policies, which is prohibitive when working with real-world mechanical plants or robots.

One way of using data efficiently (and therefore keep the number of experiments small) is to learn forward models of the underlying dynamical system, which are then used for internal simulations and policy learning. These ideas have been successfully applied to RL, control and robotics in~\cite{Schmidhuber1990,Atkeson1997c,Bagnell2001,Contardo2013,Pan2014, DeisenrothFR:2015,Pan2014,Hoof2015, Levine2015}, for instance. However, these methods use heuristic or engineered low-dimensional features, and they do not easily scale to data-efficient RL using pixel information only because even ``small'' images possess thousands of dimensions.

A common way of dealing with high-dimensional data is to learn low-dimensional feature representations. Deep learning architectures, such as deep neural networks~\cite{Hinton2006}, stacked auto-encoders~\cite{Bengio2007,Vincent2008}, or convolutional neural networks~\cite{LeCun1998}, are the current state of the art in  learning parsimonious representations of high-dimensional data. Deep learning has been successfully applied to image, text and speech data in commercial products, e.g., by Google, Amazon and Facebook.

Deep learning has been used to produce first promising results in the context of model-free RL on images: For instance,~\cite{Mnih2015} present an approach based on Deep-Q-learning, in which human-level game strategies are learned autonomously, purely based on pixel information. Moreover, \cite{Lange2012} presented an approach that learns good discrete actions to control a slot car based on raw images, employing deep architectures for finding compact low-dimensional representations.
Other examples of deep learning in the context of RL on image data include~\cite{Cuccu2011,Koutnik2013}. These approaches have in common that they try to estimate the value function from which the policy is derived. However, neither of these algorithms learns a predictive model and are, therefore, prone to data inefficiency, either requiring data collection from millions of experiments or relying on discretization and very low-dimensional feature spaces, limiting their applicability to mechanical systems.

To increase data efficiency, we therefore introduce a model-based approach to learning from pixels to torques. In particular, exploit results from~\cite{Wahlstrom2015} and jointly learn a lower-dimensional embedding of images and a transition function in this lower-dimensional space that we can use for internal simulation of the dynamical system. For this purpose, we employ deep auto-encoders for the lower-dimensional embedding and a multi-layer feed-forward neural network for the transition function. We use this deep dynamical model to predict trajectories  and apply an adaptive model-predictive-control (MPC) algorithm \cite{Mayne:2014} for online closed-loop control, which is practically based on pixel information only. 

MPC has been well explored in the control community, However, adaptive MPC has so far not received much attention in the literature \cite{Mayne:2014}. An exception is~\cite{Sha:2008}, where the authors advocate a neural network approach similar to ours. However, they do not consider high-dimensional data but assume that they have direct access to low-dimensional measurements.

Our approach benefits from the application of model-based optimal control principles within a machine learning framework. Along these lines, \cite{Deisenroth2009,Abramova2012,Boedecker2014,Pan2014,Levine2015} suggested to first learn a transition model and then use optimal control methods to solve RL problems. Unlike these methods, our approach does not need to estimate value functions and scales to high-dimensional problems.

Similar to our approach,~\cite{Boots2014, Levine2015, Hoof2015} recently proposed model-based RL methods that learn policies directly from visual information. Unlike these methods, we exploit a low-dimensional feature representation that allows for fast predictions and online control learning via MPC.

\subsection*{Problem Set-up and Objective}
We consider a classical $N$-step finite-horizon RL setting in which an agent attempts to solve a particular task by trial and error. 
In particular, our objective is to find a closed-loop policy $\pi^*$ that minimizes the long-term cost $V^\pi = \sum\nolimits_{\tind=0}^{\tend-1} f_0(x_\tind, u_\tind)$, where $f_0$ denotes an immediate cost, $x_\tind\in\R^D$ is the continuous-valued system state and $u_\tind\in\R^F$ are continuous control inputs. 

The learning agent faces the following additional challenges: (a) The agent has no access to the true state, but perceives the environment only through high-dimensional pixel information (images), (b) a good control policy is required in only a few trials. This setting is practically relevant, e.g., when the agent is a robot that is monitored by a video camera based on which the robot has to learn to solve tasks fully autonomously. 
Therefore, this setting is an instance of the pixels to torques problem.

\section{Deep Dynamical Model}
\label{sec:model}
Our approach to solve the pixels-to-torques problem is based on a deep dynamical model (DDM), which jointly (i) embeds high-dimensional images in a low-dimensional feature space via deep auto-encoders and (ii) learns a predictive forward model in this feature space~\cite{Wahlstrom2015}.
In particular, we consider a DDM with control inputs $u$ and high-dimensional observations $\meas$.  We assume that the relevant properties of $\meas$ can be compactly represented by a feature variable $\feature$. The two components of the DDM, i.e., the low-dimensional embedding and the prediction model, which predicts future observations $\meas_{\tind+1}$ based on past observations and control inputs, are detailed in the following.
Throughout this paper,  $\meas_\tind$ denotes the high-dimensional measurements, $\feature_\tind$ the corresponding low-dimensional encoded features and $\widehat \meas_\tind$ the reconstructed high-dimensional measurement. Further, $\widehat \feature_{\tind+1}$ and $\widehat \meas_{\tind+1}$ denote a predicted feature and measurement at time $t+1$, respectively.

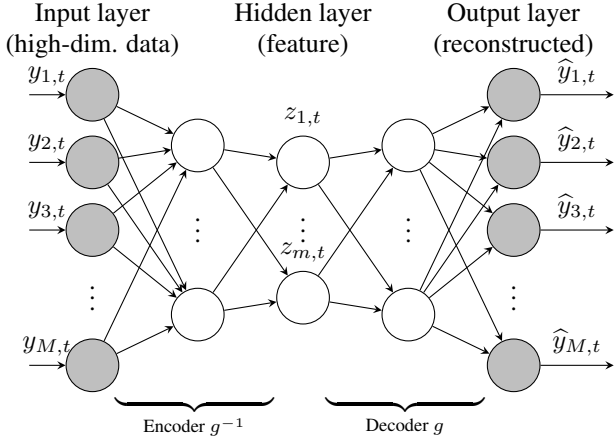
\begin{figure}
\centering
\tikzset{%
  every neuron/.style={
    circle,
    draw,
    minimum size=0.7cm    
  },
  neuron missing/.style={
    draw=none, 
    fill=none, 
    scale=1,
    text height=0.333cm,
    execute at begin node=\color{black}$\vdots$
  },
}

\begin{tikzpicture}[x=0.7cm, y=0.9cm, >=stealth]

\foreach \m/\l [count=\y] in {1,2,3,missing,4}
  \node [every neuron/.try, fill=lightgray, neuron \m/.try] (input-\m) at (0,2.5-\y) {};

\foreach \m [count=\y] in {1,missing,2}
  \node [every neuron/.try, neuron \m/.try ] (hiddena-\m) at (2,2-\y*1.25) {};

\foreach \m [count=\y] in {1,missing,2}
  \node [every neuron/.try, neuron \m/.try ] (hiddenb-\m) at (4,1.5-\y) {};
	
\foreach \m [count=\y] in {1,missing,2}
  \node [every neuron/.try, neuron \m/.try ] (hiddenc-\m) at (6,2-\y*1.25) {};
	
\foreach \m/\l [count=\y] in {1,2,3,missing,4}
  \node [every neuron/.try, fill=lightgray, neuron \m/.try] (output-\m) at (8,2.5-\y) {};

\foreach \l [count=\i] in {1,2,3,\measdim}
  \draw [<-] (input-\i) -- ++(-1.2,0)
    node [above, midway] {$\meas_{\l,\tind}$};

\foreach \l [count=\i] in {1,\featuredim}
  \node [above] at (hiddenb-\i.north) {$\feature_{\l,\tind}$};

\foreach \l [count=\i] in {1,2,3,\measdim}
  \draw [->] (output-\i) -- ++(1.9,0)
    node [above, midway] {$\widehat{\meas}_{\l,\tind}$};

\foreach \i in {1,...,4}
  \foreach \j in {1,...,2}
    \draw [->] (input-\i) -- (hiddena-\j);

\foreach \i in {1,...,2}
  \foreach \j in {1,...,2}
    \draw [->] (hiddena-\i) -- (hiddenb-\j);
\foreach \i in {1,...,2}
  \foreach \j in {1,...,2}
    \draw [->] (hiddenb-\i) -- (hiddenc-\j);
		
\foreach \i in {1,...,2}
  \foreach \j in {1,...,4}
    \draw [->] (hiddenc-\i) -- (output-\j);

\foreach \l [count=\x from 0] in {Input layer \\(high-dim. data), , Hidden layer \\ (feature), , Output layer \\ (reconstructed)}
  \node [align=center, below] at (\x*2,3) {\l}; 
	\foreach \l [count=\x from 0] in { ,$\underbrace{\qquad\qquad\qquad}_{\text{Encoder }g^{-1}}$ , , $\underbrace{\qquad\qquad\qquad}_{\text{Decoder }g}$ , }
  \node [align=center, below] at (\x*2,-2.8) {\l}; 
\end{tikzpicture}
\caption{Auto-encoder that consists of an encoder $g\inv$ and a decoder $g$. The encoder maps the original image $\meas_\tind \in\R^M$  onto its low-dimensional representation $\feature_\tind = g\inv(\meas_\tind)\in\R^m$, where $\featuredim \ll \measdim$; the decoder maps this feature back to a high-dimensional representation $\widehat{\meas}_{\tind} = g(\widehat{\feature}_{\tind})$. The gray color represents high-dimensional observations.
}
\label{fig:autoencoder}
\end{figure}

\begin{figure}[t]
\tikzset{%
  every neuron/.style = {
    circle,
    draw,
    minimum size=1.1cm
  },  
	neuron missing/.style={
		draw=none, 
		scale=1,
		text height=0.333cm,
	},
}
\centering
\begin{tikzpicture}[x=0.9cm, y=0.9cm, >=stealth]
\node [every neuron/.try] (u0) at (-4,-0.5) {};
\node [neuron missing/.try] (u0b) at (-4,-0.5) {$u_{\tind-\ndim+1}$};
\node [neuron missing/.try] (ua) at (-2.5,-0.5) {$\cdots$};
\node [every neuron/.try] (u1) at (-1,-0.5) {$u_{\tind}$};
\node [neuron missing/.try] (ya) at (-2.5,1.5) {$\cdots$};
\node [every neuron/.try] (y0) at (-4,1.5) {};
\node [neuron missing/.try] (y0b) at (-4,1.5) {$\feature_{\tind-\ndim+1}$};
\node [every neuron/.try] (y1) at (-1,1.5) {$\feature_{\tind}$};
\node [every neuron/.try] (y2) at (1,1.5) {};
\node [neuron missing/.try] (y2b) at (1,1.5) {\small{$\widehat{\feature}_{\tind+1 \mid \hn}$}};
\node [neuron missing/.try] (d1) at (-2.5,-0.5) {};
\node [neuron missing/.try] (d2) at (-2.5,1.5) {};
\node [neuron missing/.try] (d3) at (-2.5,3.3) {};
\node [every neuron/.try,fill=lightgray] (z0) at (-4,3.3) {};
\node [neuron missing/.try] (z0b) at (-4,3.3) {$\meas_{\tind-\ndim+1}$};
\node [neuron missing/.try] (za) at (-2.5,3.3) {$\cdots$};
\node [every neuron/.try,fill=lightgray] (z1) at (-1,3.3) {$\meas_{\tind}$};
\node [every neuron/.try,fill=lightgray] (z2) at (1,3.3) {};
\node [every missing/.try] (z2b) at (1,3.3) {\small{$\widehat{\meas}_{\tind+1 \mid \hn}$}};
\node [align=center, left] at (-4.7,3.3) {High-dim. \\ observations};
\node [align=center, left] at (-4.7,1.5) {Features}; 
\node [align=center, left] at (-4.7,-0.5) {Control \\ inputs};
\draw [->] (y1) -- (y2);
\draw [->] (u1) -- (y2);
\draw [->] (u0) -- (y2) node[midway,below] {$f$};
\draw [->] (z1) -- (y1) node[midway,left] {$g^{-1}$};
\draw [->] (z0) -- (y0) node[midway,left] {$g^{-1}$};
\draw [->] (y2) -- (z2) node[midway,left] {$g$};
\draw[->] (y0) .. controls (-1,0.7) .. (y2);
\end{tikzpicture}
\caption{Prediction model: Each feature $\feature_i$ is computed from high-dimensional data $\meas_i$ via the encoder $g^{-1}$. The transition model predicts the feature $\widehat{\feature}_{\tind+1 \mid \hn}$ at the next time step based  on the $\ndim$-step history of $\ndim$ past features $\feature_{\tind-\ndim+1},\dotsc,\feature_{\tind}$ and control inputs $u_{\tind-\ndim+1},\dotsc,u_{\tind}$. The predicted feature $\widehat{\feature}_{\tind+1 \mid \hn}$ can be mapped to a high-dimensional prediction $\widehat{y}_{t+1}$ via the decoder $g$. The gray color represents high-dimensional observations.}
\label{fig:model}
\end{figure}
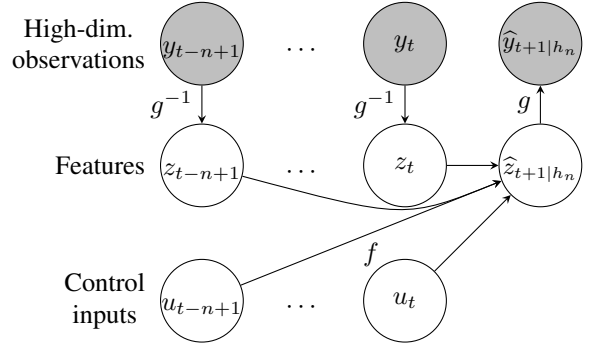

\subsection{Deep Auto-Encoder}
\label{sec:autoencoder}
We use a deep auto-encoder for embedding images in a low-dimensional feature space, where both the encoder $g^{-1}$ and the decoder~$g$ are modeled with deep neural networks.
Each layer $k$ of the \emph{encoder} neural network $g^{-1}$ computes $\meas^{(k+1)}_\tind = \sigma(A_k \meas^{(k)}_\tind + b_k)$, where $\sigma$ is a sigmoidal activation function (we used $\arctan$) and $A_k$ and $b_k$ are free parameters. The input to the first layer is the image, i.e., $\meas^{(1)}_\tind = \meas_\tind$. The last layer is the low-dimensional feature representation of the image $\feature_\tind(\theta_{\text{E}}) = g^{-1}(\meas_\tind;\theta_{\text{E}})$, where $\theta_{\text{E}} = [\dots, A_k, b_k,\dots]$ are the parameters of all neural network layers.
The \emph{decoder} $g$ consists of the same number of layers in reverse order, see \fig\ref{fig:autoencoder}, and approximately inverts the encoder $g$, such that $\widehat{\meas}_{\tind}(\theta_{\text{E}},\theta_{\text{D}}) = g(g^{-1}(\meas_\tind;\theta_{\text{E}});\theta_{\text{D}}) \approx \meas_\tind$
is the reconstructed version of $\meas_\tind$  with an associated reconstruction error
\begin{align} \label{eq:reconstruction_error}
\varepsilon_\tind^{\text{R}}(\theta_{\text{E}},\theta_{\text{D}}) = \meas_\tind - \widehat{\meas}_{\tind}(\theta_{\text{E}},\theta_{\text{D}}).
\end{align}
The main purpose of the deep auto-encoder is to keep this reconstruction error and the associated compression loss negligible, such that the features $\feature_t$ are a compact representation of the images $\meas_t$.

\subsection{Prediction Model}
\label{sec:predictor_model}

We now turn the static auto-encoder into a dynamical model that can predict future features $\widehat{\feature}_{t+1}$ and images $\widehat{y}_{t+1}$. The encoder $g\inv$ allows us to map high-dimensional observations $y_t$ onto low-dimensional features $\feature_t$. For predicting
we assume that \emph{future features} $\widehat\feature_{\tind+1\mid \hn}$ depend on an $\ndim$-step history $\hn$ of past features and control inputs, i.e., 
\begin{align} \label{eq:predictor}
\widehat{\feature}_{\tind+1 \mid \hn}(\theta_{\text{P}}) = f(\feature_{\tind},u_{\tind},\dots,\feature_{\tind-\ndim+1},u_{\tind-\ndim+1};\theta_{\text{P}}),
\end{align}
where $f$ is a nonlinear transition function, in our case a feed-forward neural network, and $\theta_{\text{P}}$ are the corresponding model parameters. This is a nonlinear autoregressive exogenous model (NARX) \cite{Ljung:1999}.
The predictive performance of the model will be important for model predictive control (see Section~\ref{sec:MPC}) and for model learning based on the prediction error~\cite{Ljung:1999}.

To predict \emph{future observations} $\widehat{\meas}_{\tind+1|\hn}$ we exploit the decoder, such that $\widehat{\meas}_{t+1|\hn} = g(\widehat{\feature}_{t+1|\hn};\theta_{\text{D}})$. The deep decoder~$g$ maps features~$\feature$ to  high-dimensional observations $\meas$ parameterized by  $\theta_{\text{D}}$. 

Now, we are ready to put the pieces together: With feature prediction model \eqref{eq:predictor} and the deep auto-encoder, the DDM predicts future features and images according to
\begin{subequations}
 \label{eq:prediction}
\begin{align}
\feature_{\tind}(\theta_{\text{E}}) & = g^{-1}(\meas_{\tind};\theta_{\text{E}}),
\\
\widehat{\feature}_{\tind+1 \mid \hn}\hspace{-0.5mm}(\theta_{\text{E}},\theta_{\text{P}})	\hspace{-0.5mm} &  = \hspace{-0.5mm} f(\feature_{\tind},u_{\tind},\dots,\feature_{\tind \unaryminus \ndim+1},u_{\tind  \unaryminus \ndim+1};\theta_{\text{P}}), \notag \\
\widehat{\meas}_{\tind+1 \mid \hn}(\theta_{\text{E}},\theta_{\text{D}},\theta_{\text{P}})
& = g(\widehat{\feature}_{\tind+1 \mid \hn}
;\theta_{\text{D}}), 
\end{align}
\end{subequations}
which is illustrated in \fig\ref{fig:model}. With this prediction model we define the  prediction error
\begin{align} \label{eq:prediction_error}
\varepsilon_{\tind+1}^{\text{P}}(\theta_{\text{E}},\theta_{\text{D}},\theta_{\text{P}}) = \meas_{\tind+1} - \widehat{\meas}_{\tind+1 \mid \hn}(\theta_{\text{E}},\theta_{\text{D}},\theta_{\text{P}}),
\end{align}
where $\meas_{\tind+1}$ is the observed image at time $\tind+1$. 


\subsection{Training}
\label{sec:training}
The DDM is parameterized by the encoder parameters $\theta_{\text{E}}$, the decoder parameters $\theta_{\text{D}}$ and the prediction model parameters $\theta_{\text{P}}$. 
In the DDM, we train both the prediction model and the deep auto-encoder jointly by finding parameters $\big(\widehat{\theta}_{\text{E}},\widehat{\theta}_{\text{D}},\widehat{\theta}_{\text{P}}\big)$, such that
\begin{subequations} 
\begin{align} 
\label{eq:joint_training}
\big(\widehat{\theta}_{\text{E}},\widehat{\theta}_{\text{D}},\widehat{\theta}_{\text{P}}\big) = &
\myargmin{\theta_{\text{E}},\theta_{\text{D}},\theta_{\text{P}}} \hspace{-0.5mm} {V_{\text{R}}(\theta_{\text{E}},\theta_{\text{D}}) 
+ V_{\text{P}}(\theta_{\text{E}},\theta_{\text{D}},\theta_{\text{P}}) },\\
V_{\text{P}}(\theta_{\text{E}}, \theta_{\text{D}},\theta_{\text{P}}) 
 & =\sum\nolimits_{\tind=1}^\tend \| \varepsilon_\tind^{\text{P}}(\theta_{\text{E}},\theta_{\text{D}},\theta_{\text{P}})\|^2,
 \label{eq:training_objective1}\\
 V_{\text{R}}(\theta_{\text{E}},\theta_{\text{D}}) 
& = \sum\nolimits_{\tind=1}^\tend \|\varepsilon_\tind^{\text{R}}
(\theta_{\text{E}},\theta_{\text{D}})\|^2
\label{eq:training_objective2},
\end{align}
\end{subequations} 
which minimizes  the sums of squared reconstruction~\eqref{eq:reconstruction_error} and  prediction~\eqref{eq:prediction_error} errors.

%
%

%
We learn all model parameters $\theta_{\text{E}}, \theta_{\text{D}}, \theta_{\text{P}}$ \emph{jointly} by solving~\eqref{eq:joint_training}.\footnote{Normally when features are used for learning dynamical models, they are first extracted from the data in a pre-processing step by minimizing~\eqref{eq:training_objective2} with respect to the auto-encoder parameters $\theta_{\text{E}}, \theta_{\text{D}}$. In a second step, the prediction model parameters $\theta_{\text{P}}$ are estimated based on these features by minimizing~\eqref{eq:training_objective1} conditioned on the estimated $\widehat \theta_{\text{E}}$ and $\widehat \theta_{\text{D}}$. 
In our experience, a problem with this approach is that the learned features might have a small reconstruction error, but this representation will not be ideal for learning a transition model. The supplementary material discusses this in more detail.}
The required gradients with respect to the parameters are computed efficiently by back-propagation, and the cost function is minimized by the BFGS algorithm~\cite{NocedalW:2006}. 
Note that in~\eqref{eq:joint_training} it is crucial to include not only the prediction error $V_{\text{P}}$, but also the reconstruction error $V_{\text{R}}$.  Without this term  the multi-step ahead prediction performance will decrease because predicted features are not consistent with features achieved from the encoder. Since we consider a control problem in this paper, multi-step ahead predictive performance is crucial. 

\paragraph{Initialization.}
With a linear activation function the auto-encoder and PCA are identical \cite{Bourlard1988}, which we exploit to initialize the parameters of the auto-encoder: The auto-encoder network is unfolded, each pair of layers in the encoder and the decoder are combined, and the corresponding PCA solution is computed for each of these pairs. We start with  high-dimensional image data at the top layer and use the principal components from that pair of layers as input to the next pair of layers. Thereby, we recursively compute a good initialization for all parameters of the auto-encoder. Similar pre-training routines are found in  \cite{Hinton2006}, in which a restricted Boltzmann machine is used instead of PCA.

In this section, we have presented a DDM that facilitates fast predictions of high-dimensional observations via a low-dimensional embedded time series.  The property of fast predictions will be exploited by the online feedback control strategy presented in the following. More details on the proposed model are given in \cite{Wahlstrom2015}.


\section{Learning Closed-Loop Policies from Images}
\label{sec:MPC}
We use the DDM for learning a closed-loop policy by means of nonlinear model predictive control (MPC). We start off by an introduction to classical MPC, before moving on to MPC on images in Section~\ref{sec:MPC on images}. 
MPC finds an optimal sequence of control signals that minimizes a $K$-step loss function, where $K$ is typically smaller than the full horizon. In general, MPC relies on (a) a reference trajectory $x_{\text{ref}}=x_1^*,\dotsc,x_K^*$ (which can be a constant reference signal) and (b) a dynamics model
\begin{align}
x_{\tind+1}  = f(x_\tind,u_\tind),
\label{eq:f}
\end{align}
which, assuming that the current state is denoted by $x_0$, can be used to compute/predict a state trajectory $\widehat  x_1,\dotsc, \widehat x_K$ for a given sequence $u_0, \dotsc, u_{K-1}$ of control signals. Using the dynamics model MPC determines an optimal (open-loop) control sequence $u_0^*,\dotsc, u_{K-1}^*$, such that the predicted trajectory $\widehat x_1,\dotsc,\widehat  x_K$ gets as close to the reference trajectory $x_\text{ref}$ as possible, such that
\begin{align} 
 \hspace{-3mm}u_0^*,\dotsc,u_{K-1}^* \in  \myargmin{ u_{0:K-1}}{ \sum_{i = 0}^{K-1} \|\widehat x_\tind- x_\tind^*\|^2 + \lambda \|u_\tind\|^2},
 \label{eq:MPC_cost_functionb}
\end{align}
where $\|\widehat x_\tind- x_\tind^*\|^2$ is a cost associated with the deviation of the predicted state trajectory $\widehat x_{0:K-1}$ from the reference trajectory $x_\text{ref}$, and $\|u_\tind\|^2$ penalizes the amplitude of the control signals.
Note that the predicted $\widehat x_\tind$ depends on all previous $u_{0:K-1}$. 
%
When the control sequence $u_0^*,\dotsc, u_{K-1}^*$ is determined, the first control $u_0^*$ is applied to the system. After observing the next state, MPC repeats the entire optimization and turns the overall policy into a closed-loop (feedback) control strategy. 

%


\subsection{MPC on Images}
\label{sec:MPC on images}
%
We now turn the classical MPC procedure into MPC on images by exploiting some convenient properties of the DDM.
The DDM allows us to predict \emph{features} $\widehat \feature_1,\dotsc,\widehat \feature_K$ based on a sequence of controls $u_0,\dotsc,u_{K-1}$. 
By comparing \eqref{eq:f} with \eqref{eq:predictor}, we define the \emph{state} $x_0$ as the present and past $\ndim-1$ features and the past $\ndim-1$ control inputs, such that
\begin{align}
x_0 = [\feature_0, \dotsc, \feature_{-\ndim+1},u_{-1},\dots,u_{-\ndim+1}].
\label{eq:state def}
\end{align}
The DDM computes the  present and past features with the encoder $\feature_{\tind} = g^{-1}(\meas_{\tind}, \theta_{\text{E}})$, such that $x_0$ is known at the current time, which matches the MPC requirement. 
Our objective is to control the system towards a desired reference image frame $\meas_{\text{ref}}$. This reference frame $\meas_{\text{ref}}$ can also be encoded to a corresponding reference feature $\feature_{\text{ref}} = g^{-1}(\meas_{\text{ref}}, \theta_{\text{E}})$, which results in the MPC objective 
\begin{align}
 u_0^*,\dotsc,u_{K-1}^* \in  \myargmin{ u_{0:K-1}}{ \hspace{-0.5mm} \sum_{\tind = 0}^{K-1} \|\widehat\feature_\tind- \feature_{\text{ref}}\|^2 \hspace{-0.5mm} +\hspace{-0.5mm}  \lambda \|u_\tind\|^2},
 \label{eq:MPC_cost_functionc}
\end{align}
where $x_0$, defined in~\eqref{eq:state def}, is the current state.
%
The gradients of the cost function \eqref{eq:MPC_cost_functionc} with respect to the control signals $u_0,\dotsc,u_{K-1}$ are computed in closed form, and we use BFGS to find the optimal sequence of control signals. Note that the objective function depends on $u_0,\dotsc,u_{K-1}$ not only via the control penalty $\|u_\tind\|^2$ but also via the feature predictions $\widehat z_{1:K-1}$ of the DDM via~\eqref{eq:predictor}.

Overall, we now have an online MPC algorithm that, given a trained DDM, works indirectly on images by exploiting their feature representation. In the following, we will now turn this into an iterative algorithm that learns predictive models from images and good controllers from scratch.

\subsection{Adaptive MPC for Learning from Scratch}
We will now turn over to describe how (adaptive) MPC can be used together with our DDM to address the pixels to torques problem and to learn from scratch.
At the core of our MPC formulation lies the DDM, which is used to predict future states~\eqref{eq:state def} from a sequence of control inputs. The quality of the MPC controller is inherently bound to the prediction quality of the dynamical model, which is typical in model-based RL~\cite{Schneider1997,Schaal1997, DeisenrothFR:2015}.

To learn models and controllers from scratch, we apply a control scheme that allows us to update the DDM as new data arrives. In particular, we use the MPC controller in an adaptive fashion to gradually improve the model by collected data in the feedback loop without any specific prior knowledge of the system at hand.
Data collection is performed in closed-loop (online MPC), and it is divided into multiple sequential trials. After each trial, we add the data of the most recent trajectory to the data set, and the model is re-trained using all data that has been collected so far. 

\begin{algorithm} 
\caption{Adaptive MPC in feature space}
\label{alg:proposed}
\begin{algorithmic} 
\STATE Follow a random control strategy and record data
\LOOP
  \STATE Update DDM with all data collected so far 
  \FOR{$t=0$ to $N-1$}
  \STATE Get state $x_t$ via auto-encoder
  \STATE $u_\tind^*\leftarrow \epsilon$-greedy MPC policy using DDM prediction
    \STATE Apply $u^*_\tind$ and record data
  \ENDFOR
\ENDLOOP
\end{algorithmic}
\end{algorithm}
Simply applying the MPC controller based on a randomly initialized model would make the closed-loop system very likely to converge to a point, which is far away from the desired reference value, due to the poor model that cannot extrapolate well to unseen states. This would in turn imply that no data is collected in unexplored regions, including the region that we actually are interested in. There are two solutions to this problem: Either we use a probabilistic dynamics model as suggested in~\cite{Schneider1997,DeisenrothFR:2015} to explicitly account for model uncertainty and the implied natural exploration or we follow an explicit exploration strategy to ensure proper excitation of the system. In this paper, we follow the latter approach. In particular, we choose an $\epsilon$-greedy exploration strategy where the optimal feedback $u_0^*$ at each time step is selected with a probability $1-\epsilon$, and a random action is selected with probability $\epsilon$.


%
Algorithm~\ref{alg:proposed} summarizes our adaptive online MPC scheme. We initialize the DDM with a random trial. We use the learned DDM to find an $\epsilon$-greedy policy using predicted features within MPC. This happens online. The collected data is added to the data set and the DDM is updated after each trial.

\section{Experimental Results}
In the following, we empirically assess the components of our proposed methodology for autonomous learning from high-dimensional synthetic image data: (a) the quality of the learned DDM and (b) the overall learning framework.

In both cases, we consider a sequence of images ($51\times 51=2601$ pixels) and a control input associated with these images.  Each pixel $\meas_\tind^{(i)}$ is a component of the measurement $\meas_\tind \in\R^{2601}$ and assumes a continuous gray-value in the interval $[0,1]$. No access to  the underlying dynamics or the state (angle $\varphi$ and angular velocity $\dot\varphi$) was available, i.e., we are dealing with a high-dimensional continuous state space. The challenge was to learn (a) a good dynamics model (b) a good controller from pixel information only. We used a sampling frequency of $0.2$\,s and a time horizon of $25$\,s, which corresponds to 100 frames per trial.

The input dimension has been reduced to $\text{dim}(\meas_\tind) = 50$ prior to model learning using PCA. With these 50-dimensional inputs, a four-layer auto-encoder network was used with dimension 50-25-12-6-2, such that the features were of dimension $\text{dim}(\feature_\tind) = 2$, which is optimal to model the periodic angle of the pendulum. The order of the dynamics was selected to be $n = 2$ (i.e., we consider two consecutive image frames) to capture velocity information, such that $z_{\tind+1} = f(z_{\tind}, u_{\tind}, z_{\tind-1}, u_{\tind-1})$. For the prediction model $f$ we used a feedforward neural network with a 6-4-2 architecture. Note that the dimension of the first layer is given by $n(\text{dim}(\feature_t) + \text{dim}(u_t)) = 2(2+1) = 6$.

\subsection{Learning Predictive Models from Pixels}
\begin{figure}[t]
\centering
	\includegraphics[width =\hsize]{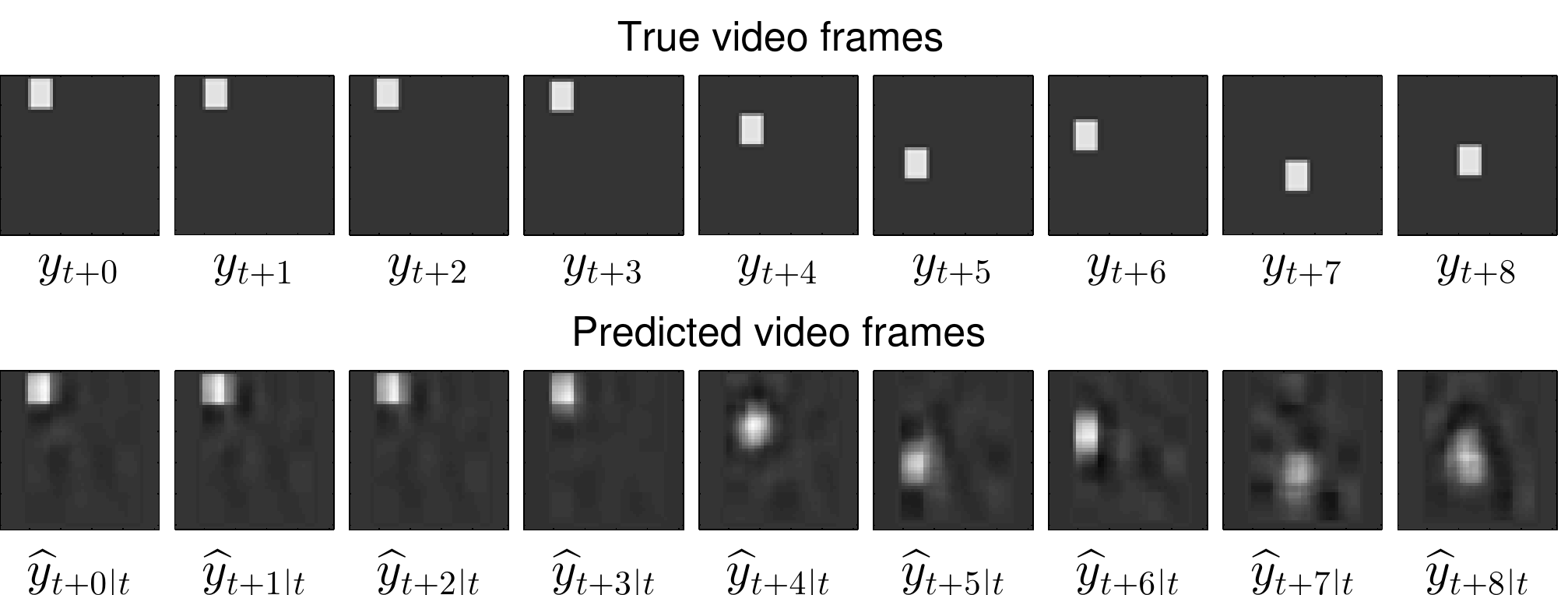}
\caption{Long-term (up to eight steps) predictive performance of the DDM: True (upper plot) and predicted (lower plot) video frames on test data.
}
\label{fig:results_long_term}
\vspace{-5mm}
\end{figure}

To assess the predictive performance of the DDM, we took $601$ screenshots of a moving tile, see \fig\ref{fig:results_long_term}. The control inputs are the (random) increments in position in horizontal and vertical directions. 

\label{sec:model_learning}
We evaluate the performance of the learned DDM in terms of long-term predictions, which play a central role in MPC for autonomous learning. Long-term predictions are obtained by concatenating multiple 1-step ahead predictions. 

\begin{figure}[h!]
\centering
\subfigure[Autoencoder and prediction model]{\includegraphics[width = 0.9\hsize]{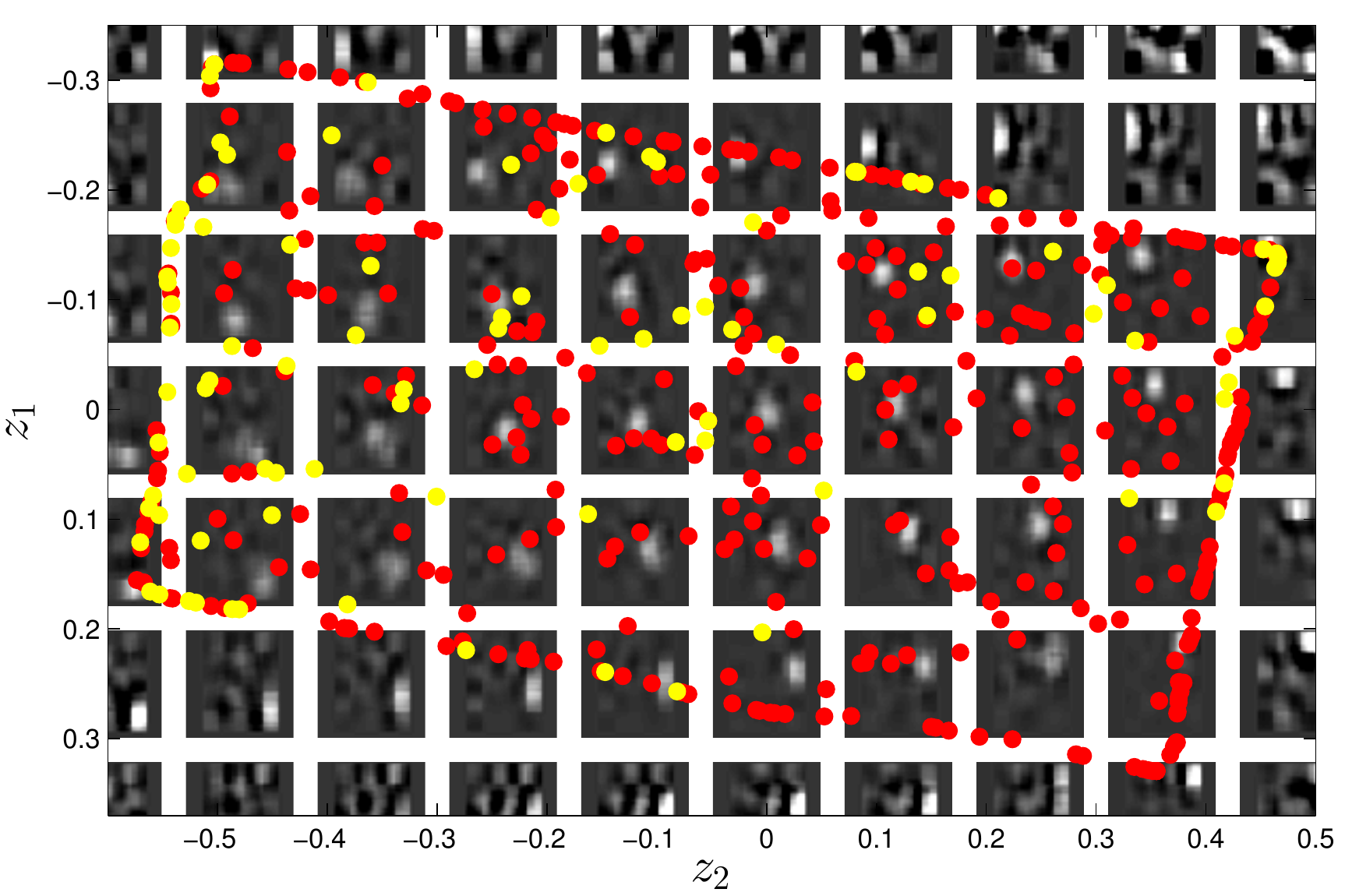}
\label{fig:latent_2dtile_new}
} \\
%
\subfigure[Only auto-encoder]{\includegraphics[width = 0.9\hsize]{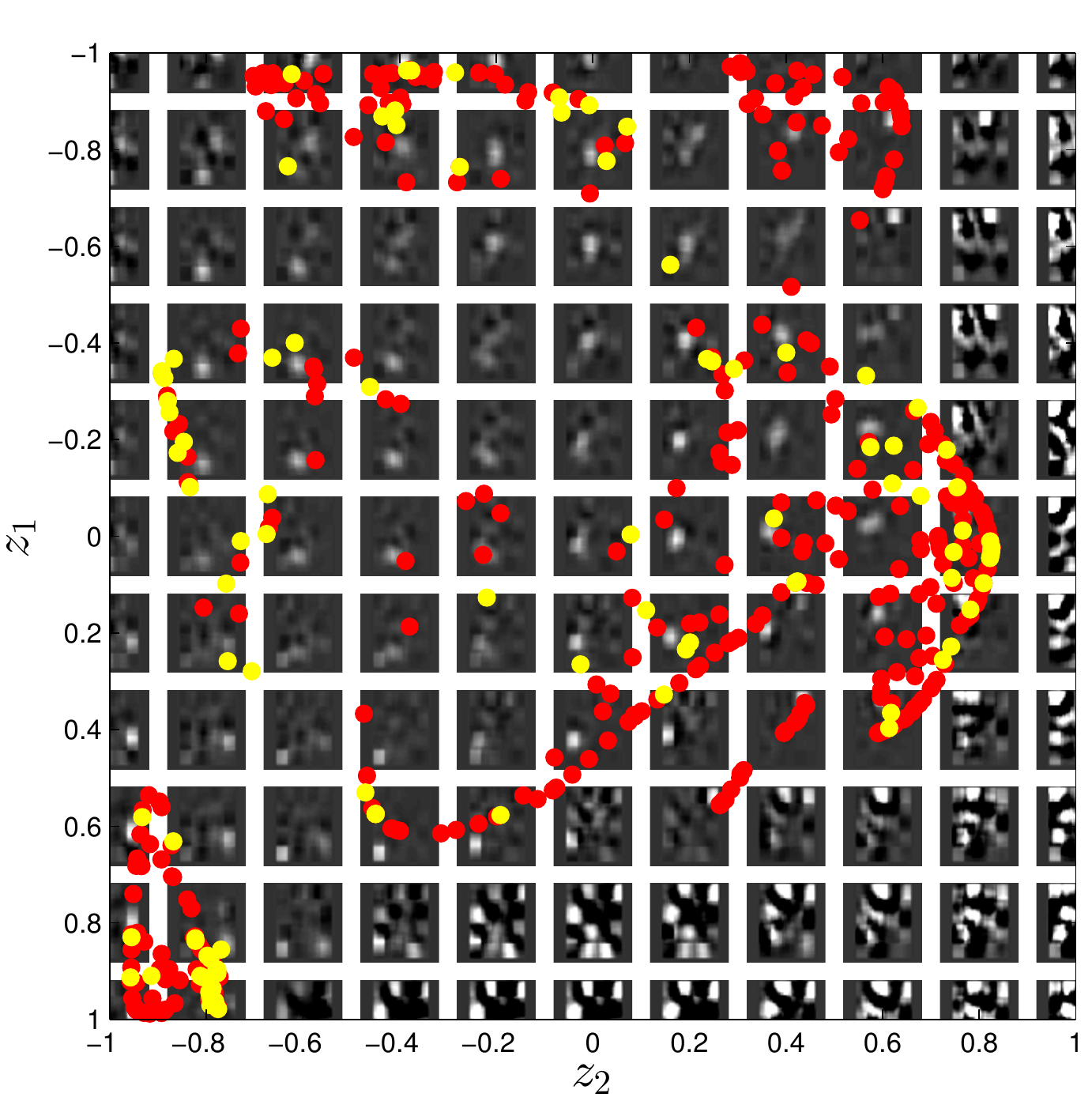}
\label{fig:latent_2dtile_sep}
}
\caption{Feature space for both joint \subref{fig:latent_2dtile_new} and sequential training \subref{fig:latent_2dtile_sep} of auto-encoder and prediction model. The feature space is divided into grid points. For each grid point the decoded high-dimensional image is displayed and the feature values for the training data (red) and validation data (yellow) are overlain. For the joint training the feature values reside on a two-dimensional manifold that corresponds to the two-dimensional position of the tile. For the separate training the feature values are scattered without structure.}
\label{fig:latent_2dtile}
\end{figure}
The performance of the DDM is illustrated in \fig\ref{fig:results_long_term} on a test data set. The top row shows the ground truth images and the bottom row shows the DDM's long-term predictions. The model predicts future frames of the tile with high accuracy both for 1-step ahead and multiple steps ahead. The model yields a good predictive performance for both one-step ahead prediction and multiple-step ahead prediction.

In \fig\ref{fig:latent_2dtile_new}, the feature representation of the data is displayed. The features reside on a two-dimensional manifold that encodes the two-dimensional position of the moving tile. By inspecting the decoded images we can see that each corner of the manifold corresponds to a corner position of the tile. Due to this structure a relatively simple prediction model is sufficient to describe the dynamics. In case the auto-encoder and the prediction model would have been learned sequentially (first training the auto-encoder, and then based on these features values train the prediction model) such a structure would not have been enforced. In \fig\ref{fig:latent_2dtile_sep} the corresponding feature representation is displayed where only the auto-encoder has been trained. Clearly, these features does not exhibit such a structure.


\subsection{Closed-Loop Policy Learning from Pixels}
\label{sec:policy_learning}
In this section, we report results on learning a policy that moves a pendulum  (1-link robot arm with length 1\,m, weight 1\,kg and friction coefficient 1\,Nsm/rad) from a start position $\varphi = 0$ to a target position $\varphi = \pm\pi$. 
The reference signal was the screenshot of the pendulum in the target position. For the MPC controller, we used a 
%
%
%
\begin{figure}[t]
\centering
	\includegraphics[width = \hsize]{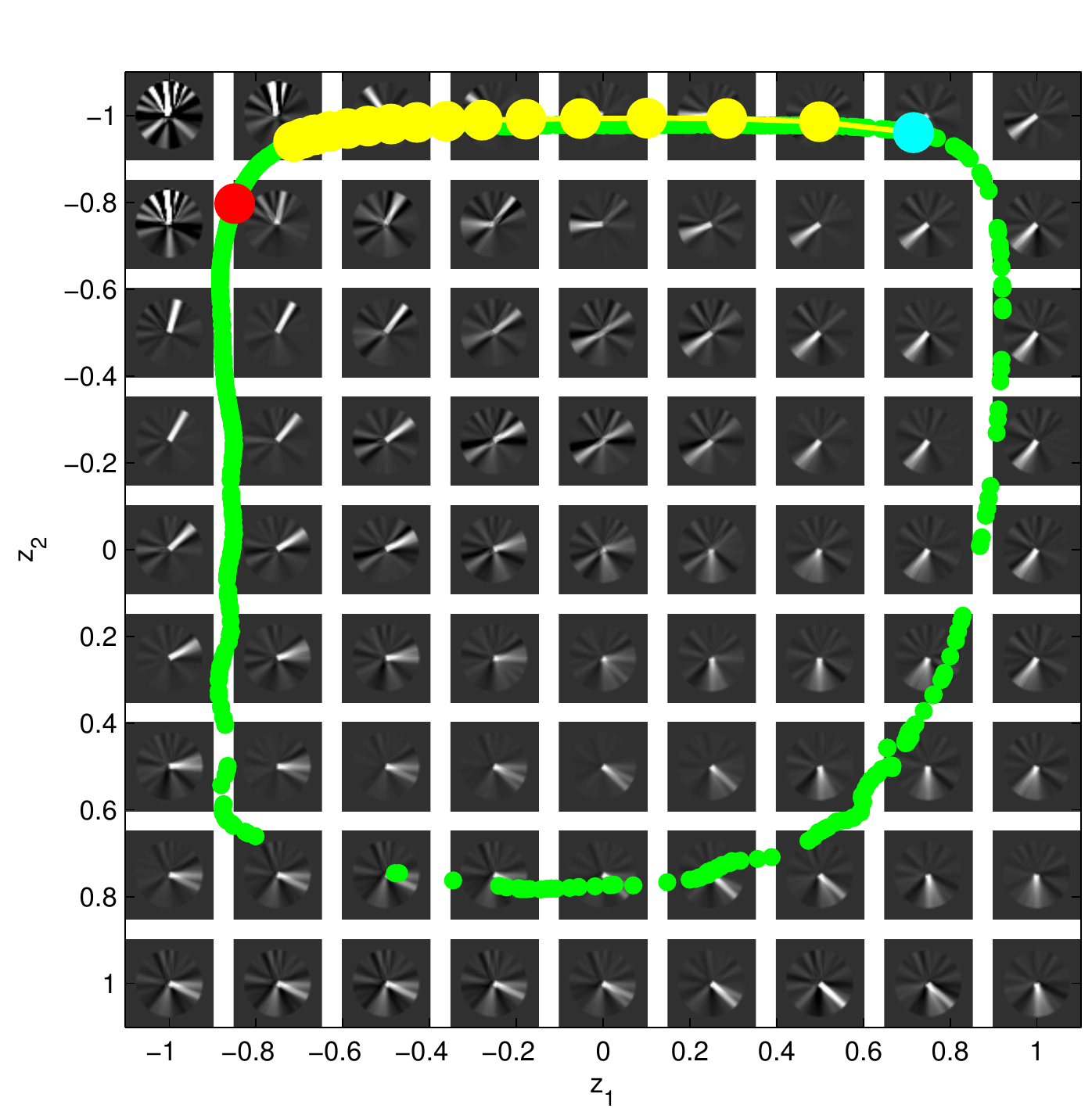}
	\caption{The feature space $\feature \in [-1, 1] \times [-1, 1]$ is divided into $9 \times 9$ grid points for illustration purposes. For each grid point the decoded high-dimensional image is displayed.  Green: Feature values that correspond to collected experience in previous trials. Cyan: Feature value that corresponds to the current time step. Red: Desired reference value. Yellow: 15-steps-ahead prediction after optimizing for the optimal control inputs.}
	\label{fig:pred_example}
\end{figure}
planning horizon of $P=15$ steps and a control penalty $\lambda=0.01$. For the $\epsilon$-greedy exploration strategy we used $\epsilon = 0.2$.
We conducted 50 independent experiments with different random initializations. The learning algorithm was run for 15 trials (plus an initial random trial). After each trial, we retrained the DDM using all collected data so far, where we also include the reference image while learning the auto-encoder.

\fig\ref{fig:pred_example} displays the decoded images corresponding to learned latent representations in $[-1,1]^2$. The learned feature values of the training data (green) line up in a circular shape, such that a relatively simple prediction model is sufficient to describe the dynamics. If we would not have optimized for both the prediction error and reconstruction error, such an advantageous structure of the feature values would not have been obtained. The DDM extracts features that can also model the \emph{dynamic} behavior compactly. The figure also shows the predictions produced by the MPC controller (yellow), starting from the current time step (cyan) and targeting the reference feature (red) where the pendulum is in the target position.

\begin{figure}
\includegraphics[width = \hsize]{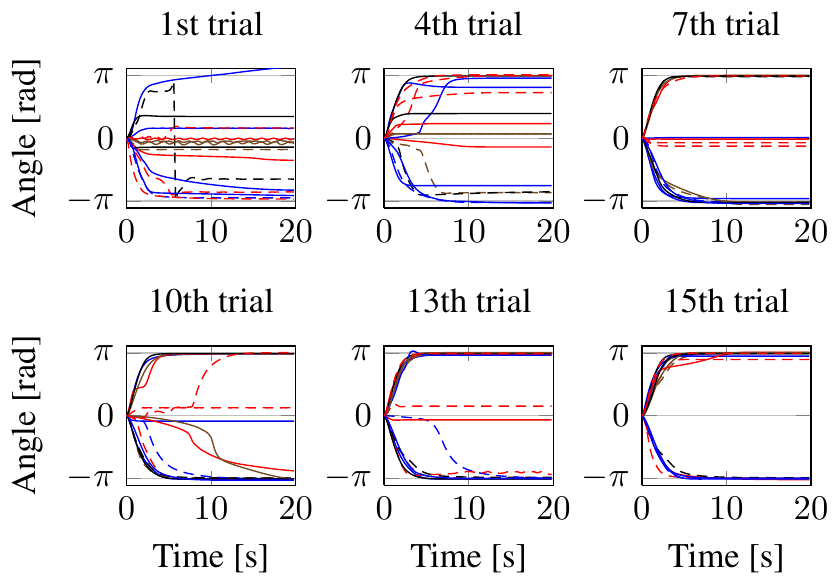}
\caption{Control performance after 1st to 15th trial evaluated with $\varepsilon = 0$ for 16 different experiments. The objective was to reach an angle of $\pm\pi$.}
\label{fig:control_traj}
\figspace
\end{figure}


%
 To assess the controller performance after each trial, we applied a greedy policy ($\epsilon = 0$). In \fig\ref{fig:control_traj}, angle trajectories for 15 of the 50 experiments at different learning stages are displayed. In the first trial, the controller managed only in a few cases to drive the pendulum toward the reference value $\pm\pi$. The control performance increased gradually with the  number of trials, and after the 15th trial, it manages in most cases to get it to an upright position.

To assess the data efficiency of our approach, we compared it with the PILCO RL framework~\cite{DeisenrothFR:2015} to learning closed-loop control policies for the pendulum task above. 
PILCO is a current state-of-the art model-based RL algorithm for data-efficient learning of control policies in continuous state-control spaces. Using collected data PILCO learns a probabilistic model of the system dynamics, implemented as a Gaussian process (GP)~\cite{Rasmussen2006}. Subsequently, this model is used to compute a distribution over trajectories and the corresponding expected cost, which is used for gradient-based optimization of the controller parameters.
%

Although PILCO uses data very efficiently, its computational demand makes its direct application impractical for many data points or high-dimensional ($\gg 20$\,D) problems, such that we had to make suitable adjustments to apply PILCO to the pixels-to-torques problem.
In particular, we performed the following experiments: (1) PILCO applied to 20D PCA features, (2) PILCO applied to 2D features learned by deep auto-encoders, (3) An optimal baseline where we applied PILCO to the standard RL setting with access to the ``true'' state ($\varphi, \dot\varphi$)~\cite{DeisenrothFR:2015}.



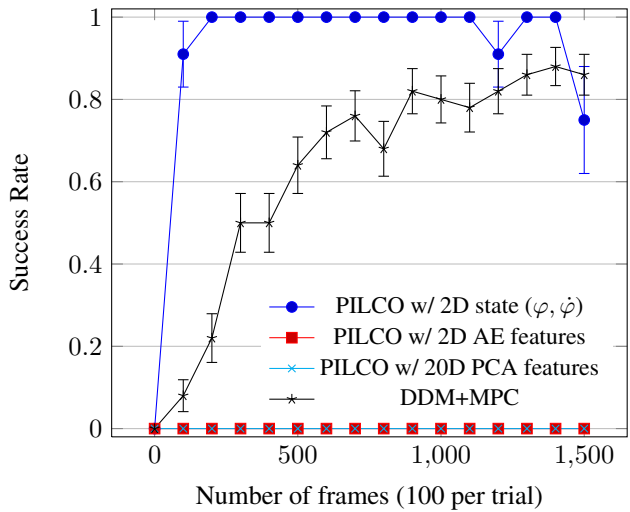
\begin{figure}[t]
\centering
\begin{tikzpicture}
\pgfsetlinewidth{40}
    \begin{axis}[
        xlabel=Number of frames (100 per trial),
        ylabel=Success Rate,
        legend style={at={(axis cs:1600,0.02)},anchor=south east, 	draw=none},
        ymajorgrids,
        ymin=-0.02,ymax=1.02,
    ] 
      \addplot+[error bars/.cd,y dir=both,y explicit]  
      coordinates {
        (0, 0) 	 +- (0,0)
        (100, 0.91) +- (0,0.08)
        (200, 1) +- (0,0)
        (300, 1) +- (0,0)
        (400, 1) +- (0,0)
        (500, 1) +- (0,0)
        (600, 1) +- (0,0)
        (700, 1) +- (0,0)
        (800, 1) +- (0,0)
        (900, 1.0) +- (0,0)
        (1000, 1) +- (0,0)
        (1100, 1.0) +- (0,0)
        (1200, 0.91) +- (0,0.08)
        (1300, 1.0) +- (0,0)
        (1400, 1.0) +- (0,0)
        (1500, 0.75) +- (0,0.13)
    };
        \addplot plot coordinates {
        (0, 0) 
        (100, 0)
        (200, 0)
        (300, 0)
        (400, 0)
        (500, 0)
        (600, 0)
        (700, 0)
        (800, 0)
        (900, 0)
        (1000, 0)
        (1100, 0)
        (1200, 0)
        (1300, 0)
        (1400, 0)
        (1500, 0)
     };  
         \addplot[color=cyan,mark=x] plot coordinates {
        (0, 0) 
        (100, 0)
        (200, 0)
        (300, 0)
        (400, 0)
        (500, 0)
        (600, 0)
        (700, 0)
        (800, 0)
        (900, 0)
        (1000, 0)
        (1100, 0)
        (1200, 0)
        (1300, 0)
        (1400, 0)
        (1500, 0)
     };  
      \addplot+[error bars/.cd,y dir=both,y explicit]  
      coordinates {
(0, 0) +- (0, 0)
(100, 0.08) +- (0, 0.038756)
(200, 0.22) +- (0, 0.059178)
(300, 0.5) +- (0, 0.071429)
(400, 0.5) +- (0, 0.071429)
(500, 0.64) +- (0, 0.068571)
(600, 0.72) +- (0, 0.064143)
(700, 0.76) +- (0, 0.061012)
(800, 0.68) +- (0, 0.066639)
(900, 0.82) +- (0, 0.054884)
(1000, 0.8) +- (0, 0.057143)
(1100, 0.78) +- (0, 0.059178)
(1200, 0.82) +- (0, 0.054884)
(1300, 0.86) +- (0, 0.04957)
(1400, 0.88) +- (0, 0.046423)
(1500, 0.86) +- (0, 0.04957)
    };
    \legend{\small PILCO w/ 2D state ($\varphi,\dot\varphi$)\\ \small PILCO w/ 2D AE features\\ \small PILCO w/ 20D PCA features\\ \small DDM+MPC  \\}
    \end{axis}
\end{tikzpicture}
\caption{Average learning success with standard errors. Blue: PILCO ground-truth RL baseline using the true state ($\varphi, \dot\varphi$). Red: PILCO with learned auto-encoder features from image pixels. Cyan: PILCO on 20D feature determined by PCA. Black: Our proposed MPC solution using the DDM.}
\label{fig:success_rate}
\end{figure}

\fig\ref{fig:success_rate} displays the average success rate of PILCO (including standard error) and our proposed method using deep dynamical models together with a tailored MPC (DDM+MPC). We define ``success'' if the pendulum's angle is stabilized within $10^\circ$ around the target state.\footnote{Since we consider a continuous setting, we have to define a target \emph{region}.}
The baseline (PILCO trained on the ground-truth 2D state ($\varphi, \dot\varphi$)) is shown in blue and solves the task very quickly.  The graph shows that our proposed algorithm (black), which learns torques directly from pixels, is not too far behind the ground-truth RL solution, achieving a n almost 90\% success rate after 15 trials (1500 image frames).
However, PILCO trained on the 2D auto-encoder features (red) and 20D PCA features fail consistently in all experiments 
We explain PILCO's failure by the fact that we trained the auto-encoder and the transition dynamics in feature space separately. The auto-encoder finds  good features that minimize the reconstruction error. However, these features are not good for modeling the dynamic behavior of the system,\footnote{When we inspected the latent-space embedding of the auto-encoder, the pendulum angles do not nicely line up along an ``easy'' manifold as in \fig\ref{fig:pred_example}. See supplementary material for more details.} and lead to bad long-term predictions.

Computation times of PILCO and our method are vastly different: While PILCO spends most time optimizing policy parameters, our model spends most of the time on learning the DDM. Computing the optimal nonparametric MPC policy happens online and does not require significant computational overhead. To put this into context, PILCO required a few days of learning time for 10 trials (in a 20D feature space). In a 2D feature space, running PILCO for 10 trials and 1000 data points requires about 10 hours.

Overall, our DDM+MPC approach to learning closed-loop policies from high-dimensional observations exploits the learned Deep Dynamical Model to learn good policies fairly data efficiently.

\section{Conclusion}
We have proposed a data-efficient model-based RL algorithm that learns closed-loop policies in continuous state and action spaces directly from pixel information. The key components of our solution are (1) a deep dynamical model (DDM) that is used for long-term predictions in a compact feature space and (2) an MPC controller that uses the predictions of the DDM to determine optimal actions on the fly without the need for value function estimation. For the success of this RL algorithm it is crucial that the DDM learns the feature mapping and the predictive model in feature space jointly to capture dynamic behavior for high-quality long-term predictions. Compared to state-of-the-art RL our algorithm learns fairly quickly, scales to high-dimensional state spaces and facilitates learning from pixels to torques.

\subsection*{Acknowledgments}
This work was supported by the Swedish Foundation for Strategic Research under the project Cooperative Localization and the Swedish Research Council under the project Probabilistic modeling of dynamical systems (Contract number: 621-2013-5524). MPD was supported by an Imperial College Junior Research Fellowship.




\begin{thebibliography}{31}
\providecommand{\natexlab}[1]{#1}
\providecommand{\url}[1]{\texttt{#1}}
\expandafter\ifx\csname urlstyle\endcsname\relax
  \providecommand{\doi}[1]{doi: #1}\else
  \providecommand{\doi}{doi: \begingroup \urlstyle{rm}\Url}\fi

\bibitem[Abramova et~al.(2012)Abramova, Dickens, Kuhn, and
  Faisal]{Abramova2012}
Abramova, Ekatarina, Dickens, Luke, Kuhn, Daniel, and Faisal, A.~Aldo.
\newblock Hierarchical, heterogeneous control using reinforcement learning.
\newblock In \emph{EWRL}, 2012.

\bibitem[Atkeson \& Schaal(1997)Atkeson and Schaal]{Atkeson1997c}
Atkeson, Christopher~G. and Schaal, S.
\newblock Learning tasks from a single demonstration.
\newblock In \emph{ICRA}, 1997.

\bibitem[Bagnell \& Schneider(2001)Bagnell and Schneider]{Bagnell2001}
Bagnell, James~A. and Schneider, Jeff~G.
\newblock Autonomous helicopter control using reinforcement learning policy
  search methods.
\newblock In \emph{ICRA}, 2001.

\bibitem[Bengio et~al.(2007)Bengio, Lamblin, Popovici, and
  Larochelle]{Bengio2007}
Bengio, Yoshua, Lamblin, Pascal, Popovici, Dan, and Larochelle, Hugo.
\newblock Greedy layer-wise training of deep networks.
\newblock In \emph{NIPS}, 2007.

\bibitem[Boedecker et~al.(2014)Boedecker, Springenberg, W\"ulfing, and
  Riedmiller]{Boedecker2014}
Boedecker, Joschka, Springenberg, Jost~Tobias, W\"ulfing, Jan, and Riedmiller,
  Martin.
\newblock Approximate real-time optimal control based on sparse {G}aussian
  process models.
\newblock In \emph{ADPRL}, 2014.

\bibitem[Boots et~al.(2014)Boots, Byravan, and Fox]{Boots2014}
Boots, Byron, Byravan, Arunkumar, and Fox, Dieter.
\newblock Learning predictive models of a depth camera \& manipulator from raw
  execution traces.
\newblock In \emph{ICRA}, 2014.

\bibitem[Bourlard \& Kamp(1988)Bourlard and Kamp]{Bourlard1988}
Bourlard, Herv{\'e} and Kamp, Yves.
\newblock Auto-association by multilayer perceptrons and singular value
  decomposition.
\newblock \emph{Biological Cybernetics}, 59\penalty0 (4-5):\penalty0 291--294,
  1988.

\bibitem[Brock(2011)]{Brock2011}
Brock, Oliver.
\newblock \emph{Berlin {Summit on Robotics: Conference Report}}, chapter Is
  {Robotics in Need of a Paradigm Shift?}, pp.\  1--10.
\newblock 2011.

\bibitem[Contardo et~al.(2013)Contardo, Denoyer, Artieres, and
  Gallinari]{Contardo2013}
Contardo, Gabriella, Denoyer, Ludovic, Artieres, Thierry, and Gallinari,
  Patrick.
\newblock Learning states representations in {POMDP}.
\newblock \emph{arXiv preprint arXiv:1312.6042}, 2013.

\bibitem[Cuccu et~al.(2011)Cuccu, Luciw, Schmidhuber, and Gomez]{Cuccu2011}
Cuccu, Giuseppe, Luciw, Matthew, Schmidhuber, J{\"u}rgen, and Gomez, Faustino.
\newblock Intrinsically motivated neuroevolution for vision-based reinforcement
  learning.
\newblock In \emph{ICDL}, 2011.

\bibitem[Deisenroth et~al.(2009)Deisenroth, Rasmussen, and
  Peters]{Deisenroth2009}
Deisenroth, Marc~P., Rasmussen, Carl~E., and Peters, Jan.
\newblock Gaussian process dynamic programming.
\newblock \emph{Neurocomputing}, 72\penalty0 (7--9):\penalty0 1508--1524, 2009.

\bibitem[Deisenroth et~al.(2015)Deisenroth, Fox, and
  Rasmussen]{DeisenrothFR:2015}
Deisenroth, Marc~P., Fox, Dieter, and Rasmussen, Carl~E.
\newblock Gaussian processes for data-efficient learning in robotics and
  control.
\newblock \emph{IEEE-TPAMI}, 37\penalty0 (2):\penalty0 408--423, 2015.

\bibitem[Hinton \& Salakhutdinov(2006)Hinton and Salakhutdinov]{Hinton2006}
Hinton, G and Salakhutdinov, R.
\newblock Reducing the dimensionality of data with neural networks.
\newblock \emph{Science}, 313:\penalty0 504--507, 2006.

\bibitem[Koutnik et~al.(2013)Koutnik, Cuccu, Schmidhuber, and
  Gomez]{Koutnik2013}
Koutnik, Jan, Cuccu, Giuseppe, Schmidhuber, J{\"u}rgen, and Gomez, Faustino.
\newblock Evolving large-scale neural networks for vision-based reinforcement
  learning.
\newblock In \emph{GECCO}, 2013.

\bibitem[Lange et~al.(2012)Lange, Riedmiller, and Voigtl\"ander]{Lange2012}
Lange, Sascha, Riedmiller, Martin, and Voigtl\"ander, Arne.
\newblock Autonomous reinforcement learning on raw visual input data in a
  real-world application.
\newblock In \emph{IJCNN}, 2012.

\bibitem[LeCun et~al.(1998)LeCun, Bottou, Bengio, and Haffner]{LeCun1998}
LeCun, Y, Bottou, L, Bengio, Y, and Haffner, P.
\newblock Gradient-based learning applied to document recognition.
\newblock \emph{Proc. of the IEEE}, 86\penalty0 (11):\penalty0 2278--2324,
  1998.

\bibitem[Levine et~al.(2015)Levine, Finn, Darrell, and Abbeel]{Levine2015}
Levine, Sergey, Finn, Chelsea, Darrell, Trevor, and Abbeel, Pieter.
\newblock End-to-end training of deep visuomotor policies.
\newblock \emph{arXiv preprint arXiv:1504.00702}, 2015.

\bibitem[Ljung(1999)]{Ljung:1999}
Ljung, L.
\newblock \emph{System Identification: Theory for the User}.
\newblock Prentice Hall, 1999.

\bibitem[Mayne(2014)]{Mayne:2014}
Mayne, David~Q.
\newblock Model predictive control: {R}ecent developments and future promise.
\newblock \emph{Automatica}, 50\penalty0 (12):\penalty0 2967--2986, 2014.

\bibitem[Mnih et~al.(2015)Mnih, Kavukcuoglu, Silver, Rusu, Veness, Bellemare,
  Graves, Riedmiller, Fidjeland, Ostrovski, and et~al.]{Mnih2015}
Mnih, Volodymyr, Kavukcuoglu, Koray, Silver, David, Rusu, Andrei~A, Veness,
  Joel, Bellemare, Marc~G, Graves, Alex, Riedmiller, Martin, Fidjeland,
  Andreas~K, Ostrovski, Georg, and et~al.
\newblock Human-level control through deep reinforcement learning.
\newblock \emph{Nature}, 518\penalty0 (7540):\penalty0 529--533, 2015.

\bibitem[Nocedal \& Wright(2006)Nocedal and Wright]{NocedalW:2006}
Nocedal, J. and Wright, S.~J.
\newblock \emph{Numerical Optimization}.
\newblock Springer, 2006.

\bibitem[Pan \& Theodorou(2014)Pan and Theodorou]{Pan2014}
Pan, Yunpeng and Theodorou, Evangelos.
\newblock Probabilistic differential dynamic programming.
\newblock In \emph{NIPS}, 2014.

\bibitem[Rasmussen \& Williams(2006)Rasmussen and Williams]{Rasmussen2006}
Rasmussen, Carl~E. and Williams, Christopher K.~I.
\newblock \emph{Gaussian {Processes for Machine Learning}}.
\newblock The MIT Press, 2006.

\bibitem[Schaal(1997)]{Schaal1997}
Schaal, Stefan.
\newblock {Learning from demonstration}.
\newblock In \emph{{NIPS}}. 1997.

\bibitem[Schmidhuber(1990)]{Schmidhuber1990}
Schmidhuber, J{\"u}rgen.
\newblock An on-line algorithm for dynamic reinforcement learning and planning
  in reactive environments.
\newblock In \emph{IJCNN}, 1990.

\bibitem[Schneider(1997)]{Schneider1997}
Schneider, Jeff~G.
\newblock Exploiting model uncertainty estimates for safe dynamic control
  learning.
\newblock In \emph{NIPS}. 1997.

\bibitem[Sha(2008)]{Sha:2008}
Sha, Daohang.
\newblock A new neural networks based adaptive model predictive control for
  unknown multiple variable non-linear systems.
\newblock \emph{IJAMS}, 1\penalty0 (2):\penalty0 146--155, 2008.

\bibitem[Sutton \& Barto(1998)Sutton and Barto]{Sutton1998}
Sutton, Richard~S. and Barto, Andrew~G.
\newblock \emph{Reinforcement {Learning: An Introduction}}.
\newblock The MIT Press, 1998.

\bibitem[van Hoof et~al.(2015)van Hoof, Peters, and Neumann]{Hoof2015}
van Hoof, Herke, Peters, Jan, and Neumann, Gerhard.
\newblock Learning of non-parametric control policies with high-dimensional
  state features.
\newblock In \emph{AISTATS}, 2015.

\bibitem[Vincent et~al.(2008)Vincent, Larochelle, Bengio, and
  Manzagol]{Vincent2008}
Vincent, P, Larochelle, H, Bengio, Y, and Manzagol, Pierre-Antoine.
\newblock Extracting and composing robust features with denoising autoencoders.
\newblock In \emph{ICML}, 2008.

\bibitem[{Wahlstr{\"o}m} et~al.(2015){Wahlstr{\"o}m}, {Sch{\"o}n}, and
  {Deisenroth}]{Wahlstrom2015}
{Wahlstr{\"o}m}, Niklas, {Sch{\"o}n}, Thomas~B., and {Deisenroth}, Marc~P.
\newblock Learning deep dynamical models from image pixels.
\newblock In \emph{SYSID}, 2015.

\end{thebibliography}

\end{document}